# Three-dimensional end-to-end deep learning for brain MRI analysis

Radhika Juglan (1), Marta Ligero (1), Zunamys I. Carrero (1), Asier Rabasco (1), Tim Lenz (1), Leo Misera (1), Gregory Patrick Veldhuizen (1), Paul Kuntke (6), Hagen H. Kitzler (6), Sven Nebelung(1,5), Daniel Truhn (5), Jakob Nikolas Kather+ (1, 2, 3, 4)

+ Correspondence to jakob-nikolas.kather@alumni.dkfz.de

1. Else Kroener Fresenius Center for Digital Health, Technical University Dresden, Dresden, Germany

2. Medical Department 1, University Hospital and Faculty of Medicine Carl Gustav Carus, Technische Universität Dresden, Dresden, Germany

3. National Center for Tumor Diseases Dresden (NCT/UCC), a partnership between DKFZ, Faculty of Medicine and University Hospital Carl Gustav Carus, TUD Dresden University of Technology, and Helmholtz-Zentrum Dresden - Rossendorf (HZDR), Dresden, Germany

4. Medical Oncology, National Center for Tumor Diseases (NCT), University Hospital Heidelberg, Heidelberg, Germany

5. Department of Diagnostic and Interventional Radiology, University Hospital Aachen, Aachen, Germany

6. Institute of Diagnostic and Interventional Neuroradiology, Faculty of Medicine and Carl Gustav Carus University Hospital, Technische Universität Dresden, Dresden, Germany

# Abstract

Deep learning (DL) methods are increasingly outperforming classical approaches in brain imaging, yet their generalizability across diverse imaging cohorts remains inadequately assessed. As age and sex are key neurobiological markers in clinical neuroscience, influencing brain structure and disease risk, this study evaluates three of the existing three-dimensional architectures, namely Simple Fully Connected Network (SFCN), DenseNet, and Shifted Window (Swin) Transformers, for age and sex prediction using T1-weighted MRI from four independent cohorts: UK Biobank (UKB, n=47,390), Dallas Lifespan Brain Study (DLBS, n=132), Parkinson's Progression Markers Initiative (PPMI, n=108 healthy controls), and Information eXtraction from Images (IXI, n=319). We found that SFCN consistently outperformed more complex architectures with AUC of 1.00 [1.00-1.00] in UKB (internal test set) and 0.85-0.91 in external test sets for sex classification. For the age prediction task, SFCN demonstrated a mean absolute error (MAE) of 2.66 (r=0.89) in UKB and 4.98-5.81 (r=0.55-0.70) across external datasets. Pairwise DeLong and Wilcoxon signed-rank tests with Bonferroni corrections confirmed SFCN's superiority over Swin Transformer across most cohorts (p<0.017, for three comparisons). No significant demographic subgroup biases were detected in either of the tasks highlighting the importance of using a diverse cohort for training DL. Explainability analysis further demonstrates the regional consistency of model attention across cohorts and specific to each task. Our findings reveal that simpler convolutional networks outperform the denser and more complex attention-based DL architectures in brain image analysis by demonstrating better generalizability across different datasets. The study emphasizes the importance of external validation and highlights the need for balanced trade-offs between model complexity and interpretability in neuroimaging applications. With growing awareness about the critical role of sex and age bias in deep learning based clinical predictions, this study offers insights into the structural and biological underpinnings of these demographic traits, thereby helping to better equip future models to recognize and mitigate such biases.

# Introduction

Brain imaging analysis using Magnetic Resonance Imaging (MRI) has been key to identifying both anatomical and functional properties related to clinical factors, enabling earlier patient diagnosis and improved treatment [1–3]. Several studies have demonstrated the importance of structural T1-weighted MRI sequences in quantifying biomarkers for neurodegenerative diseases [4–18]. These biomarkers are quantifiable characteristics derived from T1-weighted imaging studies that indicate normal biological processes, pathologic processes, or responses to therapeutic intervention. Such biomarkers include various structural measurements, such as regional brain volumes, cortical thickness, white matter integrity, and morphometric features, which can be extracted to identify disease-specific patterns.

However, previous literature has shown that patients' demographic characteristics can significantly impact the development of diagnostic tools based on anatomical imaging. For example, sex-specific neurobiological changes have been associated with several neurodegenerative [19–23] and psychosocial diseases [24–27], while the effects of brain aging on disease progression have also been widely studied [28–30]. To address these factors, several studies have sought to predict biological variables, such as sex and age, directly from brain imaging studies [31–35] to better understand their neurobiology. However, these approaches require manual delineation to extract volumetric measurements. To overcome these limitations, Deep Learning (DL) has been explored as a means to develop end-to-end predictions from MRI without the need for manual annotations [36,37].

DL has revolutionized medical image analysis by enabling automated and highly accurate predictions from complex imaging modalities, including T1-weighted brain MRI [36,38–42]. In neuroimaging, 3-dimensional (3D) model architectures, such as 3D Convolutional Neural Networks (CNNs), have been particularly effective in capturing the volumetric structure of brain MRI data, providing superior performance over 2D models for tasks that require spatial context [40,43–45]. These 3D models are crucial for applications such as brain age prediction, where spatial relationships across different brain regions play a key role [46,47]. However, the emergence of newer model architectures, such as the Shifted Window (Swin) Transformer [48], which has demonstrated superior performance in 2D medical image tasks, particularly in the field of histopathology [49,50], presents new opportunities for applying transformer-based models to 3D medical imaging. Swin Transformers, with their hierarchical structure and self-attention mechanisms, are well-suited for extracting both local and global features from images, yet their application to 3D neuroimaging tasks, such as the prediction of general clinical and biological properties remains largely unexplored.

Despite the advances in DL-based neuroimaging models for biological predictions [40,45], several key challenges persist. Many models are trained and tested on the same cohort, raising concerns about their generalizability across diverse populations. For instance, several studies have relied on training and testing within the same cohort, casting doubt on the broader applicability of their findings [36,41,44,45,51–53]. Models often overfit specific cohort characteristics, resulting in performance degradation when applied to new datasets, as highlighted in previous research [39,47].

The primary aim of this study is to systematically evaluate the implementation of 3D deep learning architectures for prediction of biological variables, specifically sex and age, directly from imaging data, while assessing their generalizability across external cohorts. Furthermore, we investigate how subgroup biases impact the model's performances. Finally, we explore robust explainability methods to better understand the decision-making processes of the developed models.

## Methods

### Data

To evaluate model generalizability, we utilized four publicly available neuroimaging datasets: UK Biobank (UKB) [54], Dallas Lifespan Brain Study (DLBS), Parkinson's Progression Markers Initiative (PPMI), and Information eXtraction from Images (IXI) (Fig. 1A, Table 1). The UKB cohort consists of high-resolution 3D T1-weighted brain MR images (Siemens Skyra 3T, MPRAGE sequence) with demographic labels for sex and age, collected across four imaging centers, with three centers used for training and one (Newcastle) held out for internal testing. The DLBS, PPMI, and IXI cohorts, used exclusively for external validation, also provide 3D T1-weighted MRI scans annotated with sex and age information. Further information on datasets can be found in Supplementary Table 1.

Furthermore, as an additional exploratory assessment of the 3D deep learning architectures for more complex clinical predictions like cardiovascular risk assessment, we make use of UKB for classification of Major Adverse Cardiovascular Events (MACE) based on cases classified under specific ICD-10 codes (I21, I22, I24, I25, I46, I63, I64, I50, I48) comprising training and test sets for both MACE detection and prediction tasks (Fig S1).

### Image Preprocessing

To standardize input data across model training and testing, preprocessing was applied to DLBS, PPMI, and IXI, as outlined in Fig. 1B. UKB images were preprocessed following standard protocols [55]. To ensure spatial consistency across images, all brain MRI scans were registered to the Montreal Neurological Institute (MNI) space using the resseg-mni library [56,57]. Brain extraction was performed using HD-BET [58] to remove non-brain structures and prevent models from learning confounding features. 3D Images were then intensity normalized (zero mean, unit variance) at the volume-level using TorchIO's ZNormalization technique, followed by center-cropping to a standardized size of (180, 180, 180) voxels[59].

### Model Architectures

We evaluated three existing 3D neural network architectures: Simple Fully Connected Network (SFCN) [37], Monai's implementation [60] of DenseNet121 [61], and Shifted Window (Swin) Transformer [48], as shown in Figure 1C. SFCN, the state-of-the-art model for brain age prediction, consists of a series of 3D convolutional layers followed by batch normalization and ReLU activations, concluding with global average pooling and fully connected layers for outputting a single age prediction or binary sex classification logits. Monai's Densenet121 is a densely connected convolutional neural network that extends connectivity across three dimensions, enhancing feature reuse by allowing each layer to receive inputs from all preceding layers. This architecture facilitates efficient gradient flow through volumetric dense blocks, improving information propagation across the network. Lastly, Monai's 3D Swin Transformer is a hierarchical transformer model that applies shifted window-based self-attention to 3D patches, enabling scalable and computationally efficient processing of volumetric data while capturing both local and global spatial dependencies across multiple resolutions.

**Training Procedures**

Models were trained on UKB using a training-validation split (2:1) with a fixed random seed (42). Hyperparameters were tuned on the validation set, and the final model performance was evaluated on the held-out test sets from different cohorts. Binary Cross-Entropy was employed as the loss function for the sex classification, while Mean Absolute Error (MAE) was utilized for regression (age prediction). A batch size of 4 and a learning rate of 10e-05 were fixed across all models. Early stopping was applied during training, halting the process if the validation loss failed to improve for 10 consecutive epochs, thereby reducing the risk of overfitting and selecting the last best model.

**Bias Correction**

To reduce systematic bias in age predictions, we applied a linear regression-based bias correction method [62]. For each model, predicted age values were regressed against their corresponding true age labels. A linear regression model was fitted to these data points to establish a relationship between the predicted and actual ages. The resulting regression coefficients were then applied to adjust the original predictions, producing bias-corrected age estimates. After correction, MAE was calculated to quantify predictive accuracy, and Pearson's correlation coefficient (r) was computed to evaluate the linear relationship between the corrected predictions and true labels.

**Statistical Methods**

Model performance was evaluated using appropriate task-specific metrics. Area Under the Receiver Operating Characteristic Curve (AUC) was used to assess sex classification performance, while MAE and Pearson's r were used to evaluate age prediction accuracy and linear association. To compare model performance, statistical significance was assessed using the pairwise DeLong tests for AUC comparisons in classification tasks and the pairwise Wilcoxon signed-rank tests for paired differences in age prediction tasks followed by Bonferroni correction of alpha.

**Subgroup Analysis**

Following model evaluation, we conducted a subgroup analysis to assess demographic-related biases in the best-performing model. The test data was stratified by age groups (for sex classification) and sex classes (for age prediction) to evaluate potential performance disparities. This analysis provided insights into any demographic distribution-based biases or limitations in model generalizability.

**Explainability**

To better understand the spatial focus of the models, we applied two complementary explainability approaches. First, attention heatmaps were generated to visualize the regions mostly responsible for driving the model's predictions for age and sex. Trained models were used for computing the gradient of the predicted output with respect to the input image, highlighting salient regions in the MRI volumes. Heatmaps from the five most confidently predicted cases in each cohort were averaged and overlaid onto a single representative brain image taken from each cohort which are visualized as 3D surface views using Mango image processing software.

Second, we conducted a volume correlation analysis by correlating model prediction scores and ground truth (GT) labels with regional gray matter volumes from 139 Image-Derived Phenotypes (IDP) provided by UKBiobank [55]. For better visualizations, these 139 IDPs were grouped into nine broader anatomical regions as done in [63] using Freesurfer lobesStrict segmentation into frontal lobe, parietal lobe, occipital lobe, temporal lobe, limbic (or cingulate) lobe, insular lobe, subcortical structures, cerebellum, and brain stem. This analysis allowed us to quantify associations between model outputs and regional brain morphology, offering insights into task-specific anatomical relevance.

# Results

**Model Training and Population Characteristics**

The models were trained on 34,918 participants from the UKBiobank (UKB) aged 40 to 71 years, with a share of 52.2% female participants. To assess cross-center generalizability, we tested the models on unseen images from a held-out center within UKB, comprising 12,472 participants aged 40 to 71 years and 54.4% female participants (Test Set 1, "internal test set"). To evaluate cross-cohort generalizability, we tested the models on three independent datasets: the Dallas Lifespan Brain Study (DLBS) (108 participants aged 40 to 70 years with 62.9% female participants; Test Set 2, external), the Parkinson's Progression Markers Initiative (PPMI) (108 healthy controls aged 40 to 70 years with 39.8% female participants; Test Set 3, external), and the Information eXtraction from Images (IXI) cohort (319 participants aged 40 to 70 years with 59.2% female participants; Test Set 4, external). The demographic characteristics of each dataset are summarized in Table 1 and Fig. 1A, while the study workflow is illustrated in Fig. 1.

**DL predicts demographic properties from 3D brain MRI across multiple cohorts**

First, we evaluate The Simple Fully Connected Network (SFCN) architecture for predicting sex and age from T1-weighted brain MR imaging studies across the UKB, PPMI, DLBS, and IXI datasets. For sex classification, SCFN achieved an Area Under the Receiver Operating Characteristic Curve (AUC) and 95% confidence interval (CI) of 1.00 [1.00 - 1.00] in UKB, 0.91 [0.86 - 0.96] in DLBS, 0.85 [0.78 - 0.92] in PPMI, and 0.87 [0.83 - 0.91] in IXI (Fig. 2, left). While the model demonstrated high predictive accuracy, its performance declined in external unseen cohorts compared to the trained UKB cohort.

For age prediction, scatter plots (Fig. 2, right) compare predicted values vs. true values. A strong correlation was observed for the UKB test set, where SFCN achieved a Mean Absolute Error (MAE) of 2.66 years and a Pearson's r of 0.89. However, performance dropped in the external datasets: DLBS (MAE = 4.98 years, r = 0.70), PPMI (MAE = 5.13 years, r = 0.59), and IXI (MAE = 5.81 years, r = 0.55). Together, these results show that 3D end-to-end deep learning can predict biological variables from brain MRI, albeit with notable performance degradation when applied across external cohorts.

As an additional exploratory analysis we evaluated Major Adverse Cardiovascular Events (MACE) detection and prediction within the UKB cohort (Fig S1), and found the AUC values of 0.57 and 0.53 respectively indicating limiting predictive performance for more complex clinical properties.

**SFCN outperforms other 3D architectures in sex classification and age prediction**

Next, we compared the SFCN model against DenseNet and SwinTransformer for both sex classification and age prediction tasks using isotropic 3D images of size 180*180*180 voxels (Fig. 2, blue, green, red). In UKB, all three models performed equally well for the sex classification, achieving an AUC of 1.00 [1.00 - 1.00]. However, in DLBS, DenseNet achieved 0.87 [0.81 - 0.93], while SwinTransformer showed a lower AUC of 0.67 [0.59 - 0.75]. In PPMI, AUCs were 0.84 [0.77 - 0.91] for DenseNet and 0.80 [0.73 - 0.88] for SwinTransformer. To compare the predictive performance of three different models, we conducted pairwise Delong tests with Bonferroni correction to account for multiple comparisons ($\alpha = 0.05/3 \approx 0.017$ for three pairwise comparisons) and found that the SwinTransformer's performance was significantly lower than SFCN (adjusted $p < 0.017$) across all cohorts except PPMI, while DenseNet did not differ significantly from SFCN on external test sets. Detailed description of the performance with Area Under Precision Recall Curves (AUPRCs) are reported in Table 2.

For age prediction, in the UKB test set, DenseNet and SwinTransformer performed worse than SFCN, with MAEs of 6.31 years and 4.86, and Pearson's r values of 0.02 and 0.60, respectively. Performance further decreased in external datasets, with DenseNet achieving an MAE of 7.33 years ($r = 0.09$) in DLBS, 6.17 years ($r = 0.11$) in PPMI, and 7.29 years ($r = 0.07$) in IXI. Similarly, SwinTransformer yielded an MAE of 7.35 years ($r = 0.03$) in DLBS, 5.85 years ($r = 0.31$) in PPMI, and 7.32 years ($r = 0.08$) in IXI. To compare the three age prediction models, we conducted pairwise Wilcoxon signed-rank tests on the three model pairs with Bonferroni correction to account for multiple corrections ($\alpha = 0.05/3 \approx 0.017$ for three pairwise comparisons). The results confirmed that SFCN significantly outperformed both DenseNet and SwinTransformer in all cohorts (adjusted $p < 0.017$). Together, these results show that SFCN consistently outperforms both Densenet and SwinTransformer architectures across all external cohorts for predicting general clinical and biological properties from 3D brain MRI. Further training characteristics for all models can be found under Supplementary Figure S4.

**Minimal bias in SFCN model performance across age and sex groups**

To assess potential biases in model performance, we evaluated the best-performing model, SFCN, across different age and sex groups. For age bias in sex classification (Fig. 3, left), SFCN maintained a high performance with an AUC of 1.00 [1.00, 1.00] across all age groups in UKB, while in DLBS, AUC values were 0.93 [0.82, 1.00], 0.90 [0.77, 1.00], 0.90 [0.75, 0.99] for age groups of 40 to 50, 50 to 60, and 60 to 70. Similarly, PPMI yielded AUCs of 0.88 [0.62, 1.00] (40 to 50 years old), 0.86 [0.74, 0.96] (50 to 60 years old), 0.83 [0.70, 0.93] (60 to 70 years old), and IXI achieved AUCs of 0.83 [0.75, 0.91] (40 to 50 years old), 0.88 [0.80, 0.94] (50 to 60 years old), 0.87 [0.80, 0.93] (60 to 70 years old). These results indicate that age did not introduce significant bias in sex classification across cohorts.

For sex bias in age prediction, we analyzed SFCN performance across male and female groups (Fig. 3, right). In UKB, the model performed slightly better for females (MAE = 2.56 years, r = 0.90) than males (MAE = 2.77 years, r = 0.89). In DLBS, the model was more accurate for males (MAE = 4.62 years, r =

0.75) than females (MAE = 4.85 years, r = 0.72). In PPMI, performance was lower for males (MAE = 5.32 years, r = 0.55) than females (MAE = 4.98 years, r = 0.62), and in IXI, the model again showed lower accuracy for males (MAE = 6.23 years, r = 0.52) than females (MAE = 5.54 years, r = 0.57). These results suggest that age prediction performance was slightly lower for males in most cohorts, though differences were not substantial, thus indicating no sex bias in age prediction.

**DL reveals consistent task-specific attention patterns for discriminative biomarkers**

To further investigate how the model makes predictions, we used heatmaps to perform a qualitative assessment of SFCN's attention patterns across the four cohorts for both tasks. As illustrated in Fig. 4, the average heatmaps of the top five most confidently predicted cases in each cohort revealed task-specific attention normalized from 0 to 1 with a threshold relevant for each cohort. Model attention was consistently distributed across the brainstem, cerebellum, central subcortical structures, and surrounding parts of the limbic lobe, insular cortex, temporal lobe, and occipital lobe for sex classification and age prediction tasks. While these findings provide insights into the model's interpretability, further validation is needed to confirm their biological and clinical significance. For a more detailed view of slice-wise attention distribution, see the 2D heatmaps in Supplementary Figures S2 and S3.

**DL enables robust correlation estimates with grey matter region volumes**

Finally, to quantitatively assess the relationship between SFCN's predictions and underlying brain structures, we analyzed correlations between grey matter region volumes and both prediction scores and ground truth (GT) labels in the UKB test set (Fig. 5). For each region, we calculated Pearson's correlation coefficients between volumes and prediction scores (r_prediction; darker shade) and then between volumes and ground truth labels (r_label; lighter shade), enabling direct comparison of model-captured versus actual neuroanatomical associations.

In sex classification, the regions demonstrating the strongest volume correlations (>0.4) with prediction scores were the Lingual Gyrus, the posterior division of the Cingulate Gyrus, and the right anterior division of the Parahippocampal Gyrus. Meanwhile, for age prediction, the Brain Stem, Ventral Striatum, and Planum Polare in the temporal lobe exhibited absolute volume correlations above 0.4, suggesting that these regions play a key role in age-related neuroanatomical changes captured by the model. For both tasks, volume correlations with prediction scores and the GT labels were observed to be consistent across all brain regions, indicating strong alignment with model predictions and anatomical variations.

# Discussion

This study explores the potential of 3D end-to-end DL for predicting demographical properties from brain MRI, specifically evaluating the generalizability of SFCN, DenseNet, and SwinTransformer across multiple external cohorts for age prediction and sex classification. While all models performed remarkably well on the internal test set (UKB) for sex classification, their performance declined when evaluated on unseen external datasets. However, SFCN demonstrated superior robustness and generalization, outperforming both DenseNet and SwinTransformer across all tasks. These results suggest that SFCN is more robust and generalizes better to external data than the other tested architectures.

The challenge of model generalization is well recognized in DL research, particularly in neuroimaging. Previous studies have attempted to predict biological variables such as age and sex from images, often using hold-out sets from the same cohort as the training data [36,44,51–53]. However, this lack of external evaluation makes it difficult to assess true model generalizability. By incorporating three independent external cohorts, our study demonstrates that external evaluation is necessary to accurately measure model performance, a conclusion supported by previous research advocating for robust validation frameworks [41,45,64]. Furthermore, it is critical to train DL models with diverse patient data to capture the variability inherent in clinical populations. This diversity not only increases the robustness of the models, but also ensures that they are better equipped to generalise across different patient demographics and conditions. By training on a large cohort like UKB, our study demonstrates that deep learning with the appropriate architectural choices might be able to learn features that are robust to variations across different demographic patterns and acquisition parameters, with further potential for testing its generalizability across even more diverse clinical groups.

Our findings also highlight the importance of architectural choices when developing generalizable brain MRI models. While more complex architectures like SwinTransformers have shown remarkable performance in computer vision tasks, their application to 3D neuroimaging data are still limited, and may require domain-specific adaptations with extensive hyperparameter tuning for optimal performance. These findings align with prior comparative studies [65], which have shown that while deeper architectures achieve state-of-the-art performance in natural image tasks, smaller models can perform comparably or better in medical imaging (e.g., 2D chest X-ray and retina images). The evidence suggests that overparameterization in deeper networks does not always improve performance, as meaningful representations often reside in the lower layers rather than deeper ones.

In addition to model generalization, our analysis examined potential biases in model performance across demographic subgroups. Many studies have previously highlighted the influence of age on sex classification performance and the role of sex differences in age prediction [63,66–69]. Our results further emphasize the need for robust validation frameworks that include diverse and independent datasets to ensure broad model applicability across different populations.

Moreover, our volume correlation analysis revealed that different brain regions contribute differently to sex classification and age prediction, suggesting that specific neuroanatomical structures carry varying levels of predictive significance for different tasks. The alignment between prediction-based and GT-volume correlations suggests that the model is learning biologically relevant morphometric patterns, rather than relying solely on image-specific artifacts. Many of these regions have been previously associated with age-related changes and sex differences [70,71], further supporting the model's ability to capture meaningful structural variations. Furthermore the general consistency of the model's attention patterns across diverse cohorts demonstrates that SFCN might be able to capture fundamental neuroanatomical features that generalize effectively across different populations and acquisition parameters.

Our additional exploratory analysis to assess the potential of 3D end-to-end deep learning for predicting more complex clinical outcomes like MACE prediction from brain MRI and its limited predictive performances suggests that while demographic properties like age and sex are tractable prediction targets for developing novel deep learning methods, inferring more complex clinical outcomes directly from structural brain MRI still remains a substantial challenge. This finding further justifies our focus on age and sex prediction as initial benchmarks for evaluating existing 3D deep learning architectures, before advancing to more complex clinical endpoints.

While this study provides valuable insights into the application of DL for neuroimaging, it is not without its limitations. First, the study was conducted using only three external cohorts of limited sample size; future research should incorporate larger and more representative datasets to further validate model robustness. Additionally, while SFCN demonstrated superior performance compared to the tested architectures with the chosen training parameters fixed for all models, future work should explore more hyperparameter tuning for the other architectures, including learning rate schedulers, weight decay, data augmentations and architectural modifications to assess whether their potential can be further unlocked for generalized 3D brain MRI analysis. Also, this study did not aim to conduct an exhaustive benchmarking of all existing 3D-DL models, and other architectures may offer further improvements [64,71]. Another limitation is that volume correlation was performed only for gray matter regions, and future work should explore potential relationships with white matter structures as they can also influence brain health and disease progression [44]. Furthermore, it remains to be seen whether regional volume differences in sex and age groups influence broader neurobiological processes and clinical outcomes. Investigating these effects may improve model interpretability and applicability in understanding disease mechanisms.

Despite the current limitations, we see a great opportunity to use this model as a basis for applications in degenerative neurological diseases such as multiple sclerosis (MS). First, it is important to investigate how neurodegeneration will affect the results of age and sex predictions, as these factors may differ in affected populations compared to healthy individuals. Understanding these relationships will help refine the accuracy and applicability of the model in clinical settings. In a subsequent phase, we envision significant potential for the model to predict disability scores or milestones of disease worsening, which serve as a critical measure of disease severity. By integrating such predictions, clinicians could gain valuable insights into the progression of disease-related neurodegeneration in diverse disease courses, ultimately improving individual patient management and treatment strategies.

## Conclusion

In conclusion, this study contributes to the ongoing development of robust and generalizable 3D-DL architectures for medical image analysis. By systematically evaluating different architectures, we identified key factors influencing predictive performance, generalizability, and interpretability. Our findings demonstrate that SFCN provides a strong balance between accuracy and efficiency, providing potential in future for automated neuroimaging analysis. While challenges remain, this study offers insights that may inform future advancements in DL for brain MRI analysis, ultimately paving the way for improved biomarker discovery and clinical decision-making in neuroimaging applications.

# Figures and Tables

## Fig 1. Study Design

a) Data Overview

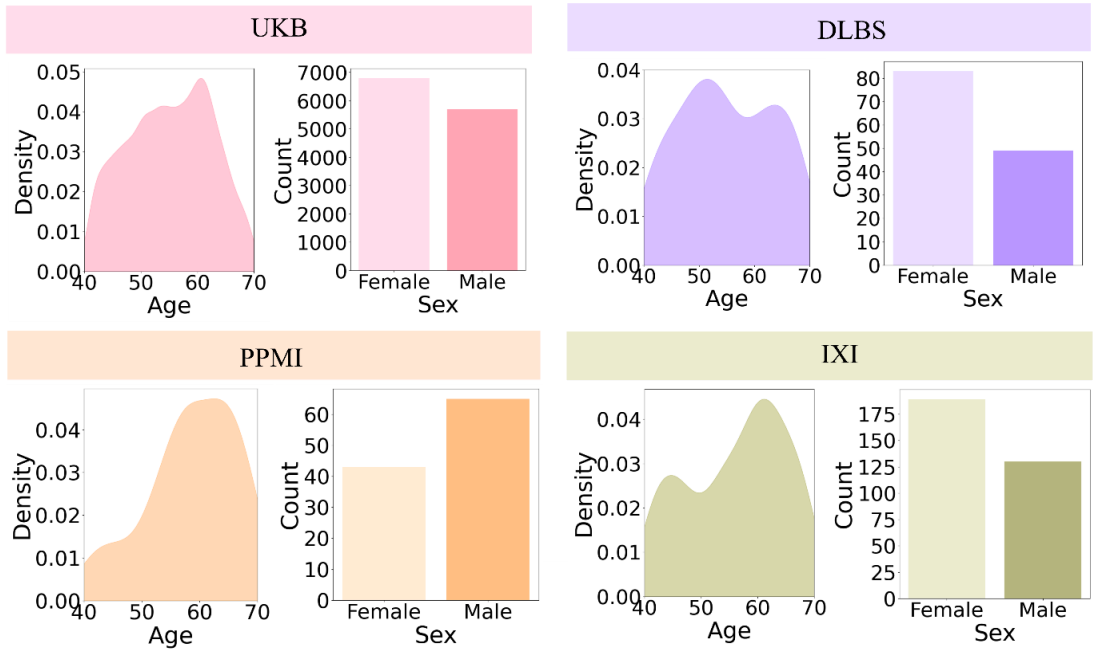

b) Image Preprocessing

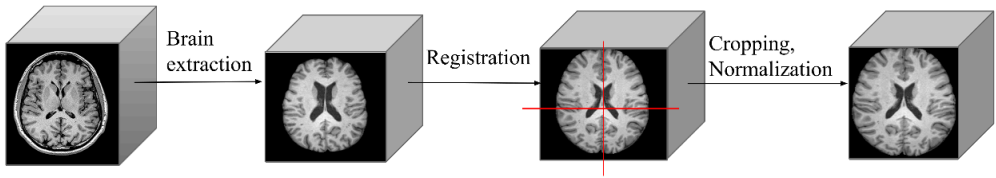

c) Modelling and Evaluation

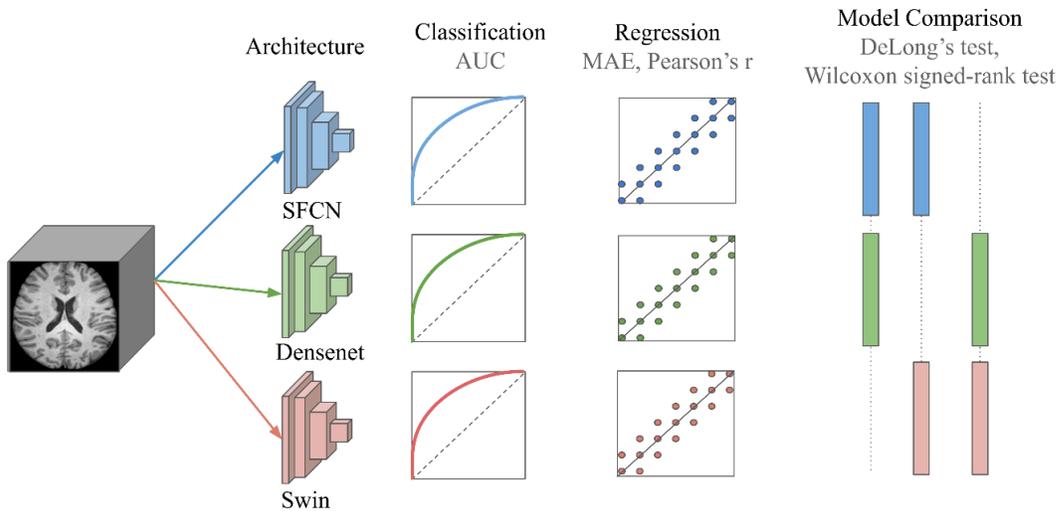

a) Four data sets with T1-weighted structural brain MR imaging studies were used: UK Biobank (UKB), Dallas Lifespan Brain Study (DLBS), Parkinson's Progression Markers Initiative (PPMI), and Information eXtraction from Images dataset (IXI). b) Images from all independent datasets went through the basic preprocessing steps of brain extraction, registration, cropping, and normalization before being fed to the 3D architectures. c) Three different 3D architectures were trained for classification and regression tasks on the images from three of the four centers from UKB cohort, while images from the Newcastle center were held out as an independent test set. The best-trained models for each task were evaluated first on the Newcastle hold-out test set from UKB, and subsequently on three additional independent cohorts to assess generalizability. The tests were done using the metric of AUCs for sex classification and MAE and Pearson's r for age prediction. Finally, the models were compared for statistical significance using the DeLongs test for the classification task and the Wilcoxon Signed-Rank test for the regression task.

**Fig 2. Model Comparison**

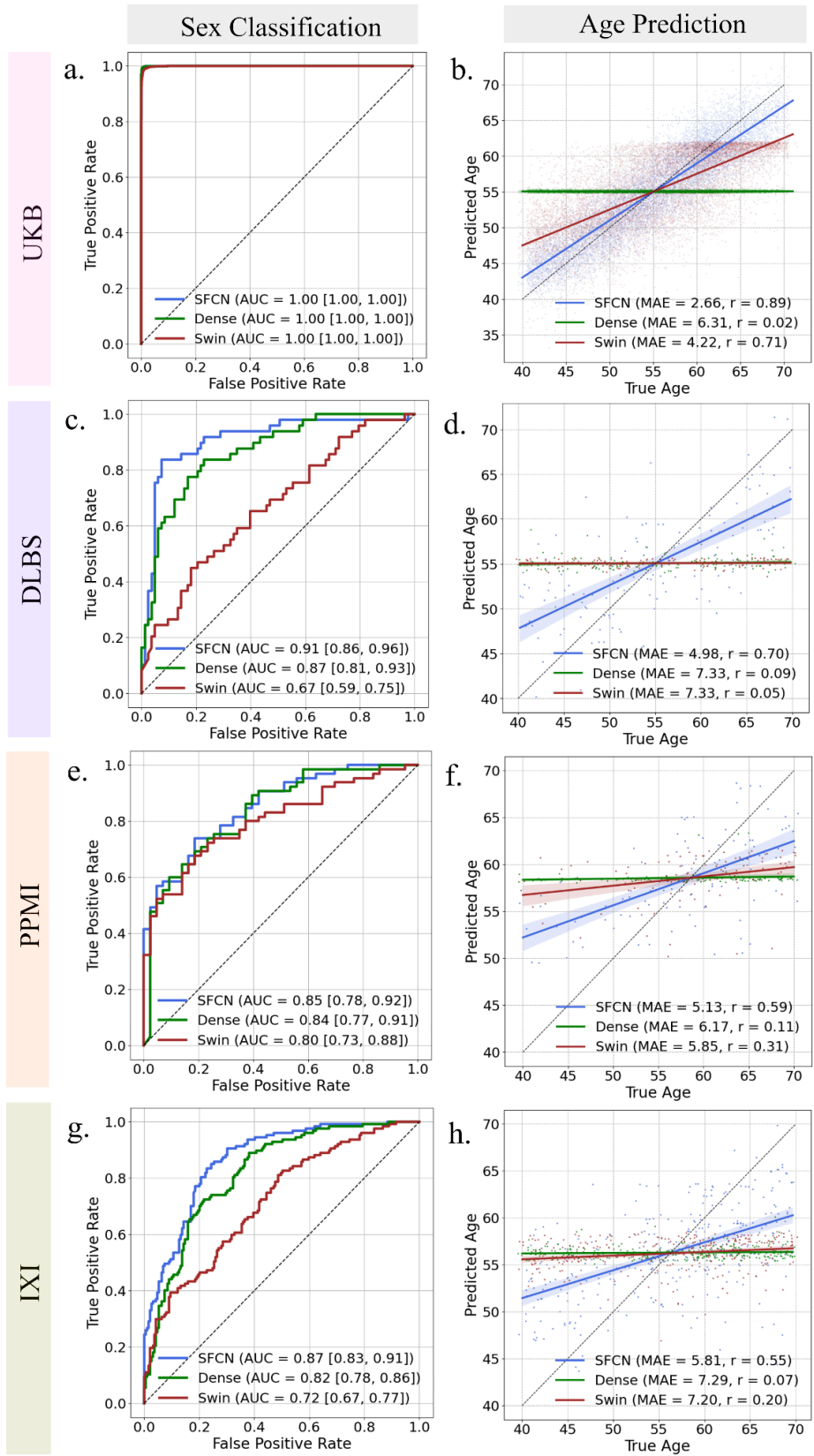

Performance comparison of three deep learning architectures - SFCN (Simple Fully Connected Network, shown in blue), DenseNet (shown in green), and Swin Transformers (shown in red) - evaluated across four neuroimaging datasets. The left column (panels a,c,e,g) shows ROC curves for sex classification performance with Area Under Receiver Operating Characteristic Curve (AUC) metrics and confidence intervals, while the right column (panels b,d,f,h) displays scatter plots for age prediction accuracy with Mean Absolute Error (MAE) and Pearson's correlation coefficient (r) metrics. Results are shown for UK Biobank (UKB, panels a-b), Dallas Lifespan Brain Study (DLBS, panels c-d), Parkinson's Progression Markers Initiative (PPMI, panels e-f), and Information eXtraction from Images dataset (IXI, panels g-h). The diagonal dotted lines in the ROC curves represent random chance performance, while in the scatter plots, they represent perfect prediction.

# Fig 3. Bias Analysis of the SFCN Model

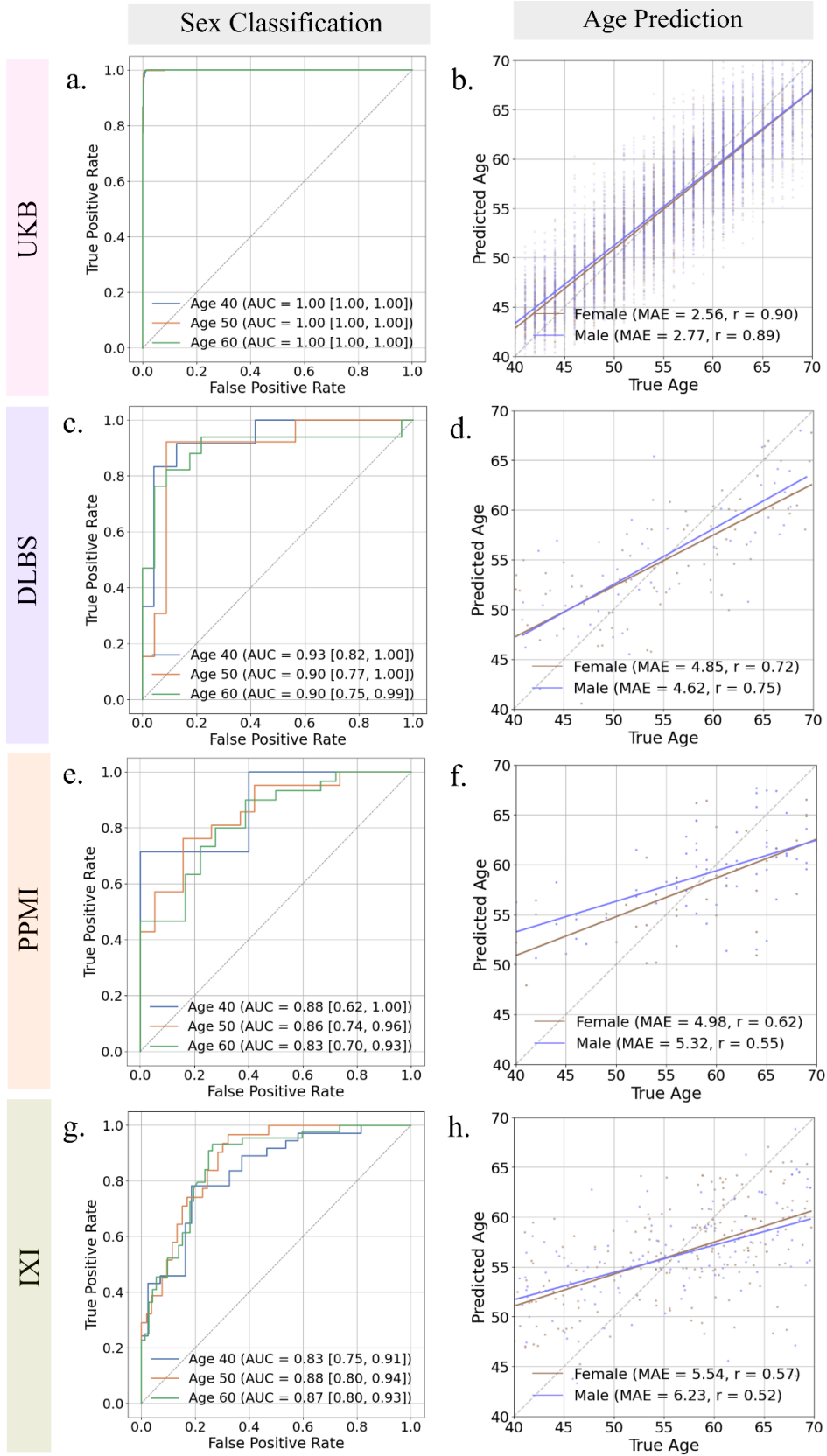

Analysis of potential age and sex biases in the SFCN (Simple Fully Connected Network) model across four neuroimaging datasets. The left column (panels a,c,e,g) evaluates sex classification performance across three age groups (40-50 years shown in blue, 50-60 years in orange, and 60-70 years in green) using ROC curves with Area Under Receiver Operating Characteristic Curve (AUC) metrics and confidence intervals. The right column (panels b,d,f,h) examines age prediction accuracy separately for females (brown) and males (violet), displaying Mean Absolute Error (MAE) and Pearson's correlation coefficients (r). Results are presented for UK Biobank (UKB, panels a-b), Dallas Lifespan Brain Study (DLBS, panels c-d), Parkinson's Progression Markers Initiative (PPMI, panels e-f), and Information eXtraction from Images dataset (IXI, panels g-h). Diagonal dotted lines represent random chance performance in ROC curves and perfect prediction in age prediction scatter plots.

**Fig 4. Average Heatmaps**

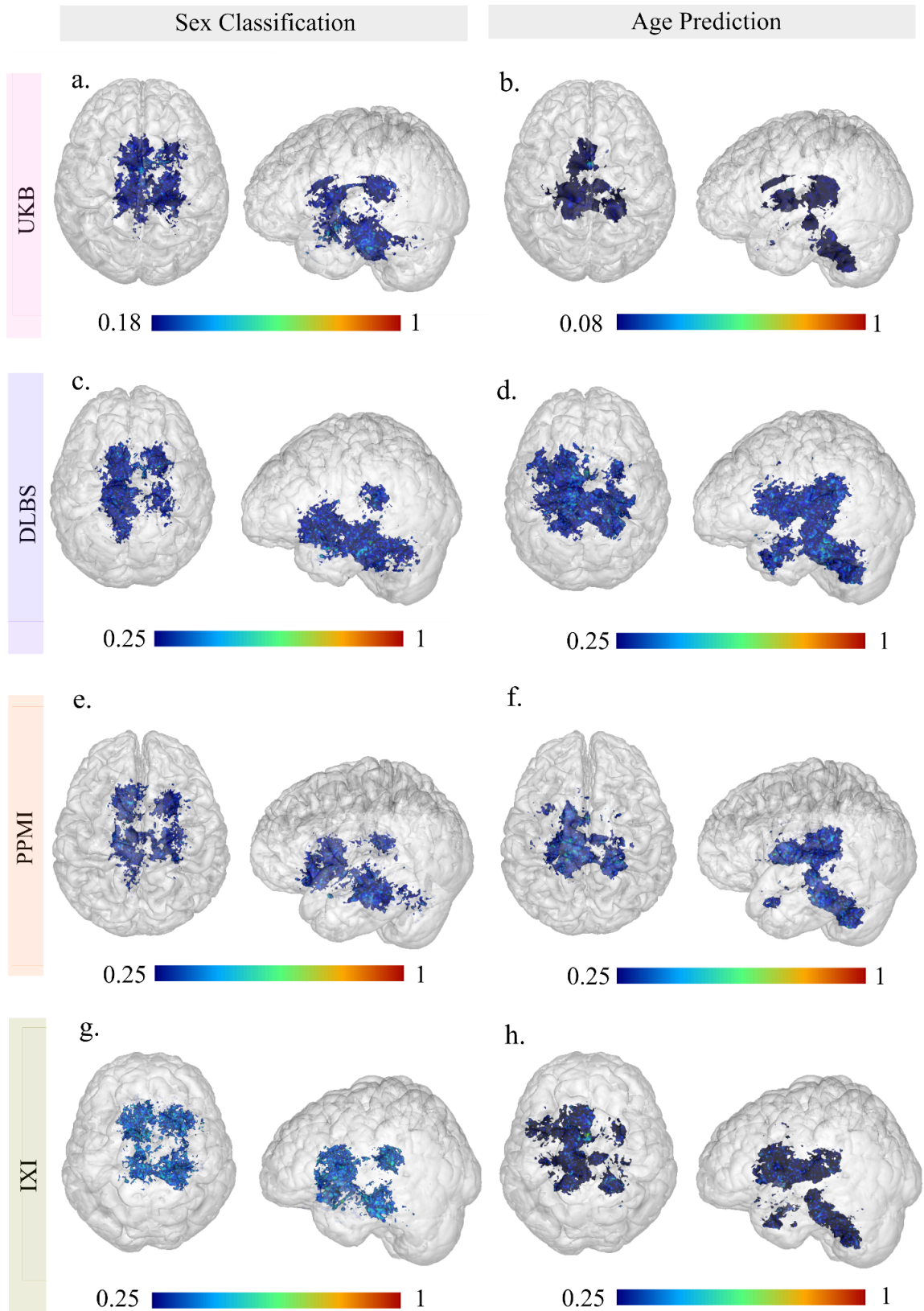

Visualization of the SFCN (Simple Fully Connected Network) model's attention patterns across brain regions using 3D glass brain representations with heatmap overlays. Each panel presents axial (top-down) and sagittal (left-side) views, where the blue-green-red spectrum represents the intensity of model attention, averaged across the top five predictions. The left column (panels a, c, e, g) depicts attention patterns for sex classification, while the right column (panels b, d, f, h) shows patterns for age prediction. Results are displayed for four neuroimaging datasets: UK Biobank (UKB, panels a–b), Dallas Lifespan Brain Study (DLBS, panels c–d), Parkinson's Progression Markers Initiative (PPMI, panels e–f), and Information eXtraction from Images (IXI, panels g–h). Attention coefficients were computed using Gradient-weighted Class Activation Mapping (Grad-CAM). The values were derived by computing the element-wise product of the input image and its gradients with respect to the predicted class, followed by normalization of the absolute values. The color bars at the bottom indicate the range of these normalized values, spanning from the minimum threshold (0.08–0.25) to the maximum (1.0).

## Fig 5. Correlations Plots

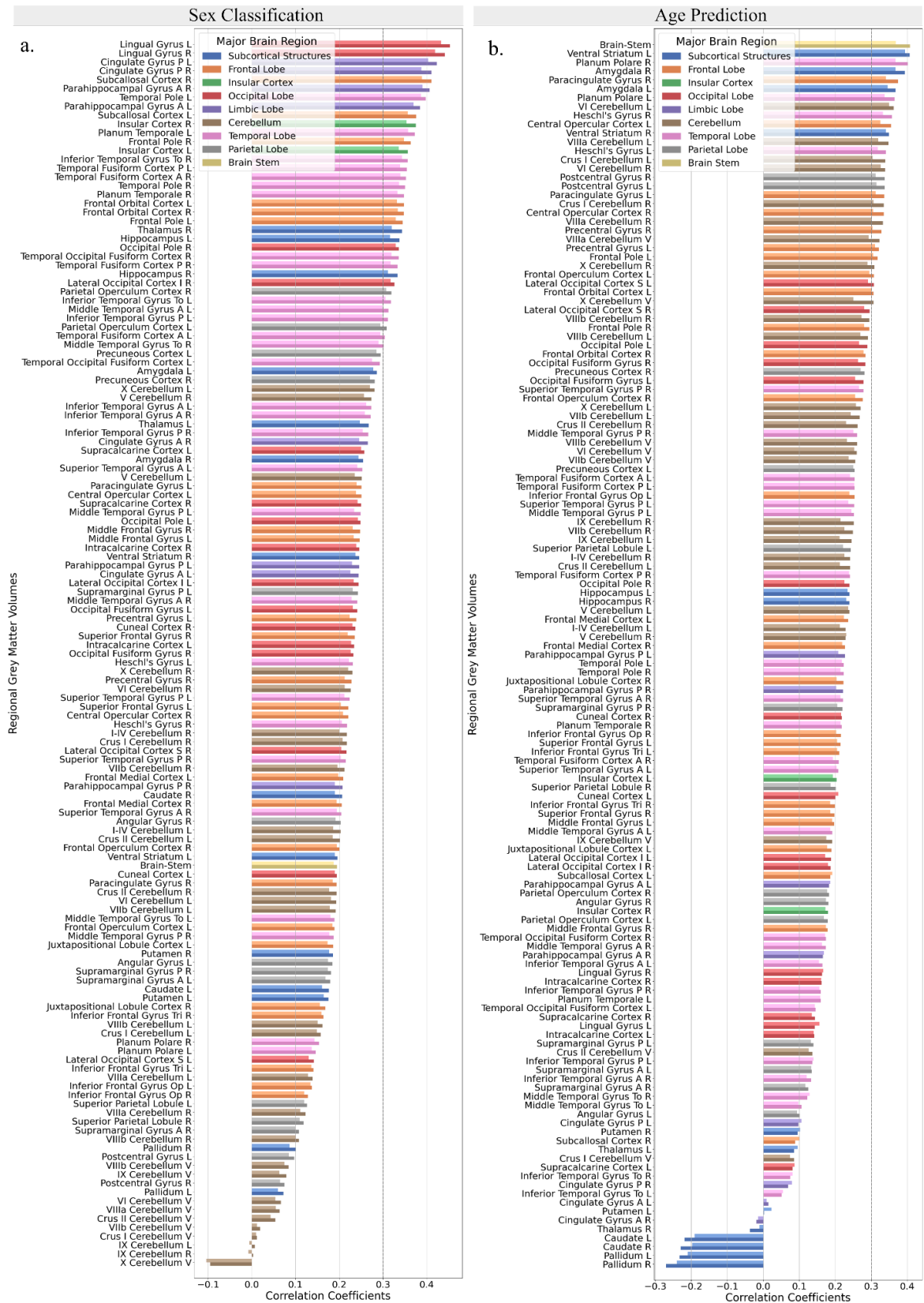

Correlations between 139 gray matter region volumes and sex/age characteristics. (a) Sex classification: Pearson's correlation coefficients of regional brain volumes with ground truth sex labels (lighter shade) and model-predicted scores (darker shade). (b) Age prediction: Pearson's correlation coefficients of regional volumes with chronological age (lighter shade) and model-predicted age scores (darker shade). Brain regions are grouped into nine major anatomical categories: subcortical structures (blue), frontal lobe (orange), insular cortex (green), occipital lobe (red), limbic lobe (purple), cerebellum (brown), temporal lobe (pink), parietal lobe (gray), and brain-stem (olive). Abbreviations were assigned to hemispheric divisions (L: Left, R: Right), anatomical subdivisions (A: Anterior, P: Posterior, S: Superior, I: Inferior), and specific locations (V: Vermis, To: Temporo-occipital part, Tri: Pars triangularis, Op: Pars opercularis). Correlation coefficients range from -0.1 to 0.4 for sex classification and -0.3 to 0.3 for age prediction.

**Table 1. Population Characteristics**

| Cohort | Center | Type | Subjects | Sex distribution | Age distribution | % female | Age range |
|---|---|---|---|---|---|---|---|
| UKB | Cheadle | | | | | | |
| | Reading | | | | | | |
| | Bristol | Train | 34,918 | 18215/16703 | 55.38 ± 7.60 | 52.17% | [40,71] |
| | Newcastle | Test set 1 | 12,472 | 6784/5688 | 55.09 ± 7.45 | 54.40% | [40,71] |
| DLBS | n.a. | Test set 2 | 132 | 83/49 | 55.08 ± 8.59 | 62.88% | [40,70] |
| PPMI | n.a. | Test set 3 | 108 | 43/65 | 58.58 ± 7.77 | 39.81% | [40,70] |
| IXI | n.a. | Test set 4 | 319 | 189/130 | 56.18 ± 8.61 | 59.25% | [40,70] |

Demographic characteristics of the study cohorts. The datasets include UK Biobank (UKB) with data from multiple centers (training set: n=34,918; test set 1; n=12,472), and three external validation cohorts: Dallas Lifespan Brain Study (DLBS, test set 2; n=132), Parkinson's Progression Markers Initiative (PPMI, test set 3; n=108), and Information eXtraction from Images (IXI, test set 4; n=319). The sex distribution is shown as female/male counts, with age presented as mean ± standard deviation. All cohorts span a common age range of 40-70 years. Center indicates specific data collection site; Type specifies dataset partition; Subjects shows total participant count; Sex distribution presents female/male numbers; Age distribution shows mean ± standard deviation; % female indicates percentage of female participants; Age range shows the minimum and maximum age in years.

**Table 2. Performance measures**

| Cohort | Model | AUC (95% CI) | AUPRC (95% CI) | MAE (95% CI) | Pearson r (95% CI) |
|---|---|---|---|---|---|
| UKB | SFCN | 0.9997 (0.9996-0.9998)* | 0.9996 (0.9995-0.9998)§ | **2.66 (2.63-2.69)** | **0.89 (0.89-0.90)** |
| | Dense | **0.9999 (0.9998-0.9999)** | **0.9998 (0.9998-0.9999)** | 6.32 (6.24-6.38)*** | 0.02 (-0.00-0.03) |
| | Swin | 0.9993 (0.9991-0.9995)*** | 0.9992 (0.9990-0.9994)*** | 4.22 (4.16-4.28)*** | 0.71 (0.70-0.72) |
| DLBS | **SFCN** | **0.91 (0.85-0.96)** | **0.86 (0.76-0.94)** | **4.97 (4.35-5.55)** | **0.70 (0.61-0.77)** |
| | Dense | 0.87 (0.80-0.92)§ | 0.80 (0.70-0.90)§ | 7.34 (6.52-8.08)*** | 0.10 (-0.10-0.30) |
| | Swin | 0.67 (0.58-0.76)*** | 0.59 (0.45-0.71)*** | 7.34 (6.57-8.11)*** | 0.05 (-0.14-0.24) |
| PPMI | **SFCN** | **0.85 (0.77-0.92)** | **0.91 (0.85-0.95)** | **5.12 (4.44-5.87)** | **0.58 (0.46-0.69)** |
| | Dense | 0.84 (0.76-0.91)§ | 0.87 (0.76-0.95)§ | 6.20 (5.34-7.11)*** | 0.11 (-0.01-0.20) |
| | Swin | 0.80 (0.72-0.88)§ | 0.88 (0.81-0.94)§ | 5.82 (5.04-6.64)* | 0.32 (0.14-0.50) |
| IXI | **SFCN** | **0.87 (0.83-0.90)** | **0.82 (0.76-0.87)** | **5.80 (5.36-6.29)** | **0.55 (0.45-0.62)** |
| | Dense | 0.82 (0.78-0.86)* | 0.73 (0.65-0.80)** | 7.29 (6.78-7.80)*** | 0.07 (-0.05-0.19) |
| | Swin | 0.72 (0.66-0.77)*** | 0.66 (0.58-0.74)*** | 7.16 (6.73-7.64)*** | 0.20 (0.10-0.30) |

The table presents the performance measures of three deep learning models (SFCN, DenseNet, and SwinTransformer) across four datasets (UKB, DLBS, PPMI, and IXI) for sex classification and age prediction tasks. The evaluation metrics include AUC and AUPRC with 95% confidence intervals for sex classification, and MAE and Pearson's correlation coefficient for the age prediction task. SFCN consistently demonstrates strong performance across all cohorts, particularly in MAE and Pearson's r for age prediction. Densenet shows competitive AUC and AUPRC in UKB but exhibits lower performance in other datasets. Statistical tests were done using the DeLong test for AUCs and AUPRCs, and the Wilcoxon signed-rank test for MAE and p-values were evaluated. Statistical significance markers represented with § for p > 0.017, * for p < 0.017, ** for p < 0.003, and *** for p < 0.0003 indicate the degree of underperformance of a model relative to the best-performing model for each metric within a given cohort.

# Additional Information

## Author contributions

RJ designed the study with the assistance from ML; RJ got access to the data with assistance from GPV; RJ analyzed the data with assistance from AR; RJ performed the experiments with the technical assistance from TL and LM; RJ wrote the manuscript with assistance from ML and ZIC; RJ got neurology specific feedback from PK and HK; RJ received radiology specific feedback from SN and DT; RJ received supervision and funding from JK. All authors provided feedback on the manuscript and collectively made the decision to submit for publication.

## Ethics Statement

The overall analysis was approved by the Ethics board at University Hospital Carl Gustav Carus, Dresden, Germany. This study adhered to the tenets of the Declaration of Helsinki.

## Consent to Participate

This study uses only anonymised secondary data from publicly available neuroimaging datasets. No new data were collected specifically for this study. All datasets were obtained in accordance with the respective data usage agreements and under appropriate ethical approvals. All scans were acquired for studies approved by local institutional review boards, research ethics committees, or human investigation committees. This research was conducted using the UK Biobank Resource under Application Number 92261. UK Biobank received ethical approval from the North West Multi-centre Research Ethics Committee (REC reference 11/NW/0382), and all participants provided informed consent. PPMI also ensures that all participants provide informed consent in accordance with IRB-approved protocols outlined in the PPMI Operations Manual and the PPMI clinical protocol that can be found at https://www.ppmi-info.org/.

**Acknowledgement**

This work is partly supported by BMBF (Federal Ministry of Education and Research) in DAAD project 57616814 (SECAI, School of Embedded Composite AI, https://secai.org/) as part of the program Konrad Zuse Schools of Excellence in Artificial Intelligence.

This research has been conducted using the UK Biobank Resource under Application Number 92261.

Data was provided in part by IXI, accessed from http://brain-development.org/ixi-dataset/.

Data used in the preparation of this article were obtained from the Parkinson's Progression Markers Initiative (PPMI) database (www.ppmi-info.org/data). For up-to-date information on the study, visit www.ppmi-info.org. PPMI—a public-private partnership—is funded by the Michael J. Fox Foundation for Parkinson's Research and funding partners; the full names of all of the PPMI funding partners can be found at www.ppmi-info.org/fundingpartners.

**Funding**

JNK is supported by the German Cancer Aid (DECADE, 70115166), the German Federal Ministry of Education and Research (PEARL, 01KD2104C; CAMINO, 01EO2101; TRANSFORM LIVER, 031L0312A; TANGERINE, 01KT2302 through ERA-NET Transcan; Come2Data, 16DKZ2044A; DEEP-HCC, 031L0315A; DECIPHER-M, 01KD2420A; NextBIG, 01ZU2402A), the German Academic Exchange Service (SECAI, 57616814), the German Federal Joint Committee (TransplantKI, 01VSF21048), the European Union's Horizon Europe research and innovation programme (ODELIA, 101057091; GENIAL, 101096312), the European Research Council (ERC; NADIR, 101114631), the National Institutes of Health (EPICO, R01 CA263318) and the National Institute for Health and Care Research (NIHR, NIHR203331) Leeds Biomedical Research Centre. The views expressed are those of the author(s) and not necessarily those of the NHS, the NIHR or the Department of Health and Social Care. This work was funded by the European Union. Views and opinions expressed are however those of the author(s) only and do not necessarily reflect those of the European Union. Neither the European Union nor the granting authority can be held responsible for them. LM is funded by NUM 2.0 (FKZ: 01KX2121). RJ is funded by the German Academic Exchange Service (SECAI, 57616814).

**Conflicts of interest**

JNK declares consulting services for Bioptimus; Panakeia; AstraZeneca; and MultiplexDx. Furthermore, he holds shares in StratifAI, Synagen, Tremont AI and Ignition Labs; has received an institutional research grant by GSK; and has received honoraria by AstraZeneca, Bayer, Daiichi Sankyo, Eisai, Janssen, Merck, MSD, BMS, Roche, Pfizer, and Fresenius.
DT received honoraria for lectures by Bayer, GE, Roche, AstraZeneca, and Philips and holds shares in StratifAI GmbH, Germany and in Synagen GmbH, Germany.

**Data availability**

This study utilized MRI data from the UK Biobank under Application Number 92261. Researchers can request access to the UK Biobank data through the official application process (https://www.ukbiobank.ac.uk/enable-your-research/apply-for-access). External validation was conducted using data from the Dallas Lifespan Brain Study (DLBS) that can be accessed via https://fcon_1000.projects.nitrc.org/indi/retro/dlbs.html or https://openneuro.org/datasets/ds004856/versions/1.0.0, Parkinson's Progression Markers Initiative (PPMI) that can be accessed via https://www.ppmi-info.org/, and Information eXtraction from Images (IXI), all of which can be accessed via https://brain-development.org/ixi-dataset/.

**Code availability**

The underlying code for this study is available on Github and can be accessed via this link https://github.com/jrad9921/RadBrainDL. Codebase is continuously developed and might evolve after publication.

# Supplementary Material

## Table S1. Dataset Information

| Dataset | Full cohort name | MRI Type | MRI Sequence | Source |
|---|---|---|---|---|
| UKB | UK Biobank | T1w | MPRAGE | https://biobank.ctsu.ox.ac.uk/crystal/crystal/docs/brain_mri.pdf |
| DLBS | Dallas Lifespan Brain Study | T1w | MPRAGE | https://fcon_1000.projects.nitrc.org/indi/retro/dlbs.html |
| PPMI | Parkinson's Progression Markers Initiative | T1w | MPRAGE | https://www.ppmi-info.org/ |
| IXI | Information eXtraction from Images | T1w | n.a. | https://brain-development.org/ixi-dataset/ |

Links for finding details about image acquisition protocol followed by each cohort.

**Fig S1. MACE prediction and detection using the SFCN model**

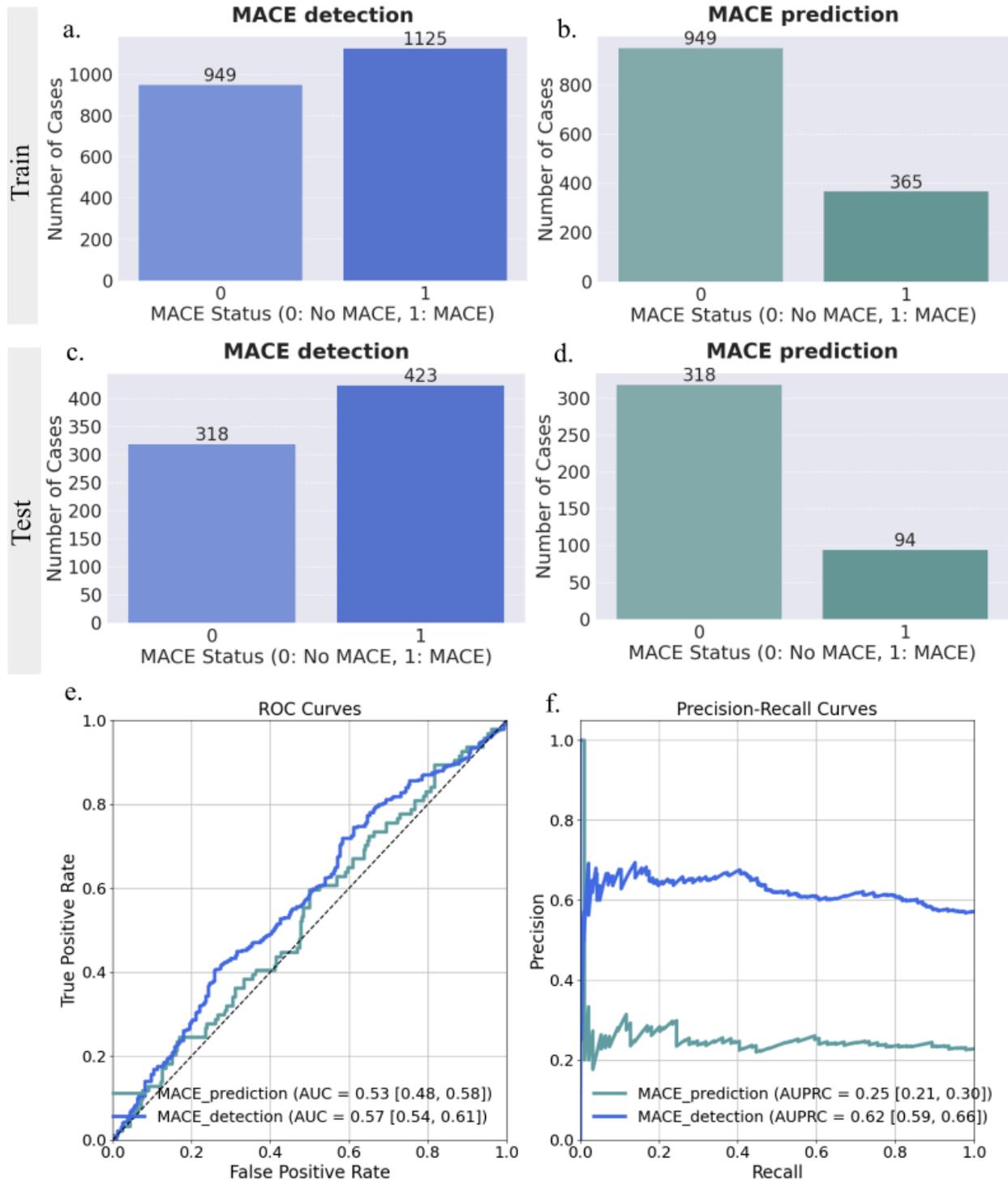

Performance of MACE (Major Adverse Cardiovascular Events) detection and prediction using the SFCN model. Panels a-d show the distribution of cases in both training and test sets: the training set contains 949 non-MACE and 1125 MACE cases for detection (a), and 949 non-MACE and 365 MACE cases for prediction (b). The test set comprises 318 non-MACE and 423 MACE cases for detection (c), and 318 non-MACE and 94 MACE cases for prediction (d). Performance metrics are presented in panels e-f, with ROC curves (e) showing AUC values of 0.57 (95% CI: 0.54-0.61) for MACE detection and 0.53 (95% CI: 0.48-0.58) for MACE prediction. Precision-Recall curves (f) demonstrate AUPRC values of 0.62 (95% CI: 0.59-0.66) for detection and 0.25 (95% CI: 0.21-0.30) for prediction, reflecting the model's performance in identifying MACE classified under specific ICD10 codes (I21, I22, I24, I25, I46, I63, I64, I50, I48).

**Fig S2. 2D Heatmaps for sex classification**

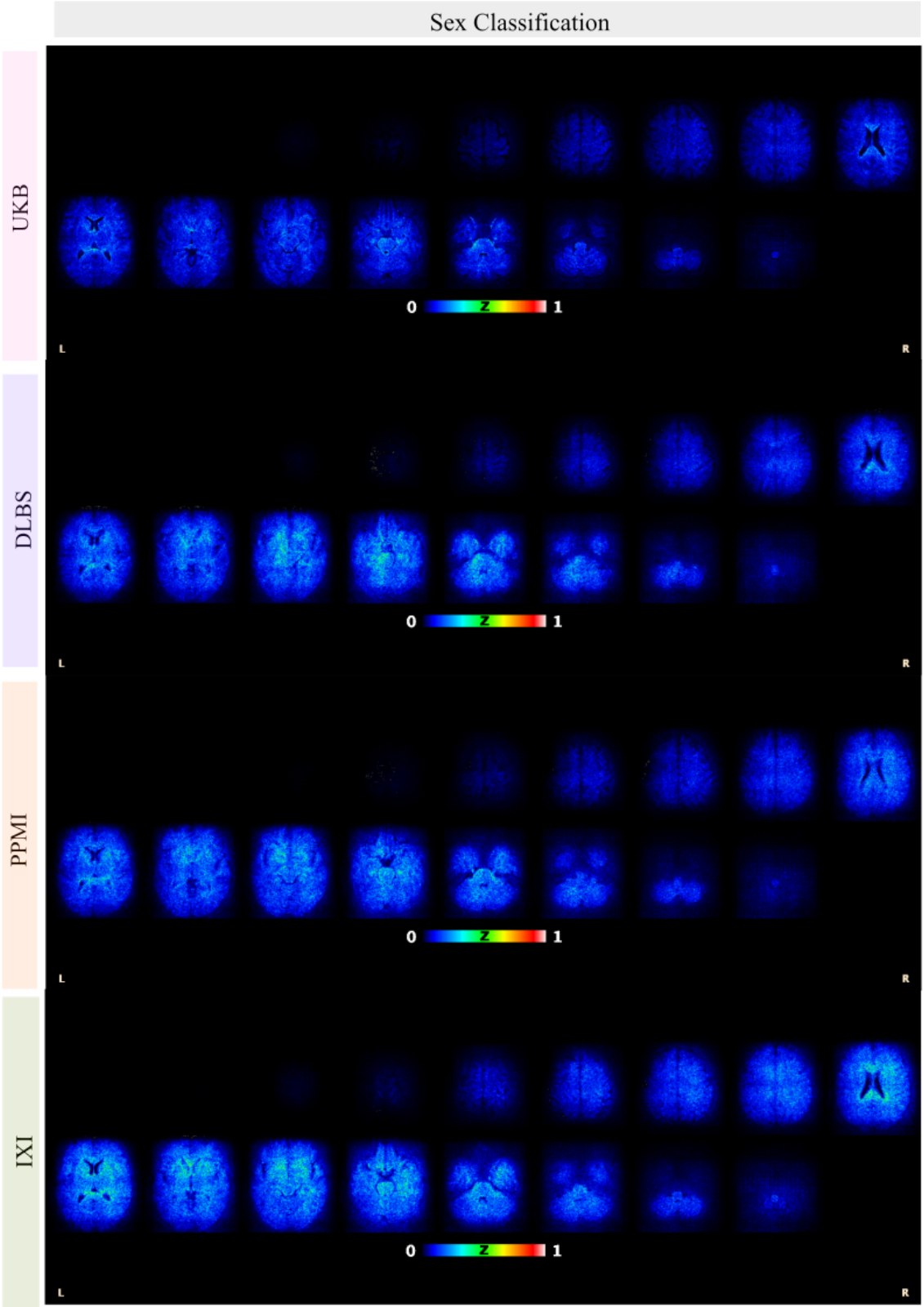

Axial brain slices in gaps of 10, demonstrating sex classification patterns for the Simple Fully Connected Network (SFCN) in four neuroimaging datasets (UKB, DLBS, PPMI, and IXI). Eighteen axial slices from top to bottom of the brain are represented in two rows of nine slices for each dataset. Left (L) and right (R) orientation markers are provided for each row, and a standardized colorbar appears beneath each dataset's slices. The intensity values are displayed on a scale of 0 to 1, represented by a blue-green-red colormap, with predominant activations appearing in blue.

**Fig S3. 2D Heatmaps for age prediction**

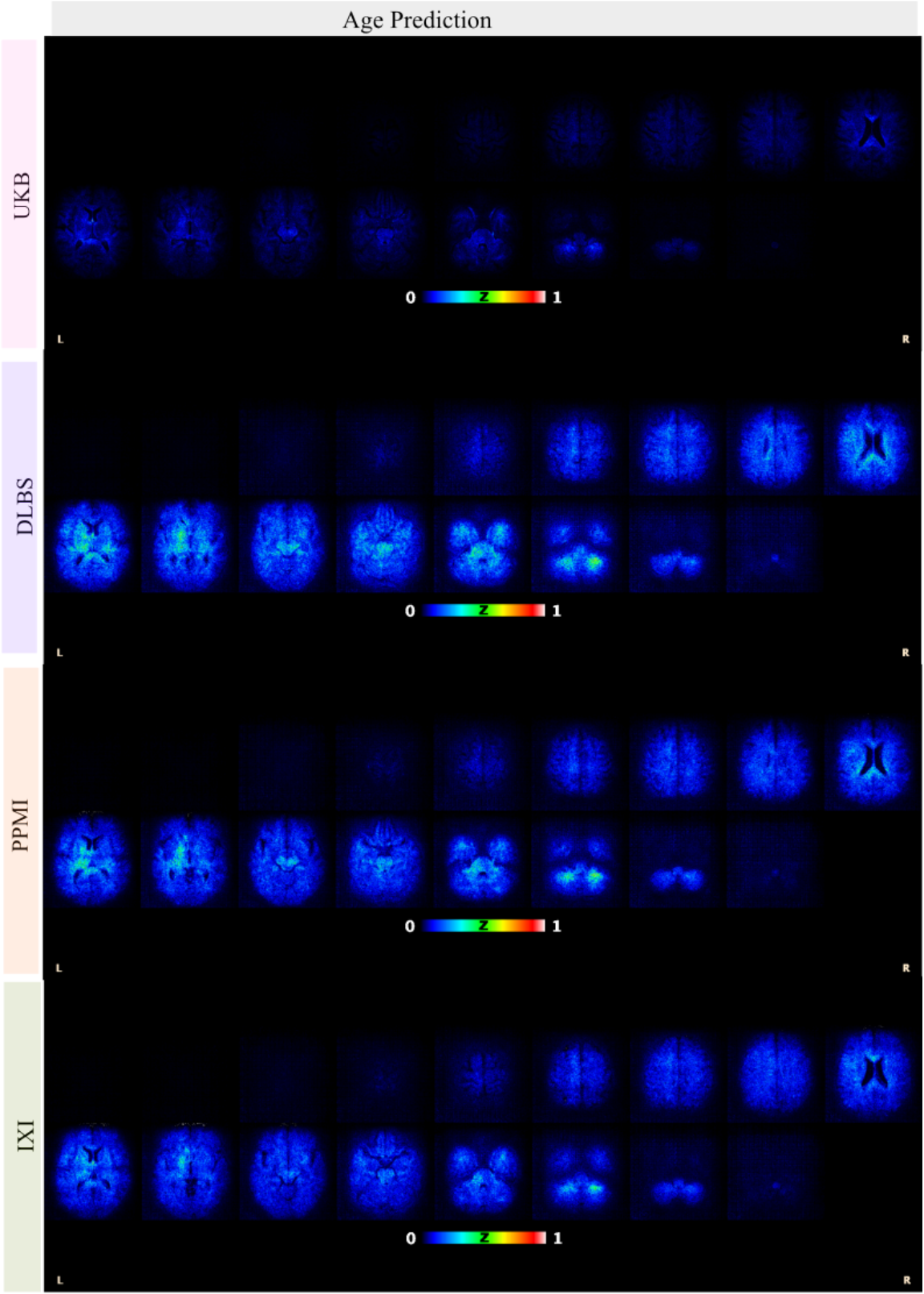

Axial brain slices in gaps of 10, demonstrating age prediction patterns for the Simple Fully Connected Network (SFCN) in four neuroimaging datasets (UKB, DLBS, PPMI, and IXI). Eighteen axial slices progressing from top to bottom of the brain are represented in two rows of nine slices for each dataset. Left (L) and right (R) orientation markers are provided for each row, and a standardized colorbar appears beneath each dataset's slices. The intensity values are displayed on a scale of 0 to 1, represented by a blue-green-red colormap, with predominant activations appearing in blue.

**Fig S4. Training and Validation Losses for each model across both cohorts**

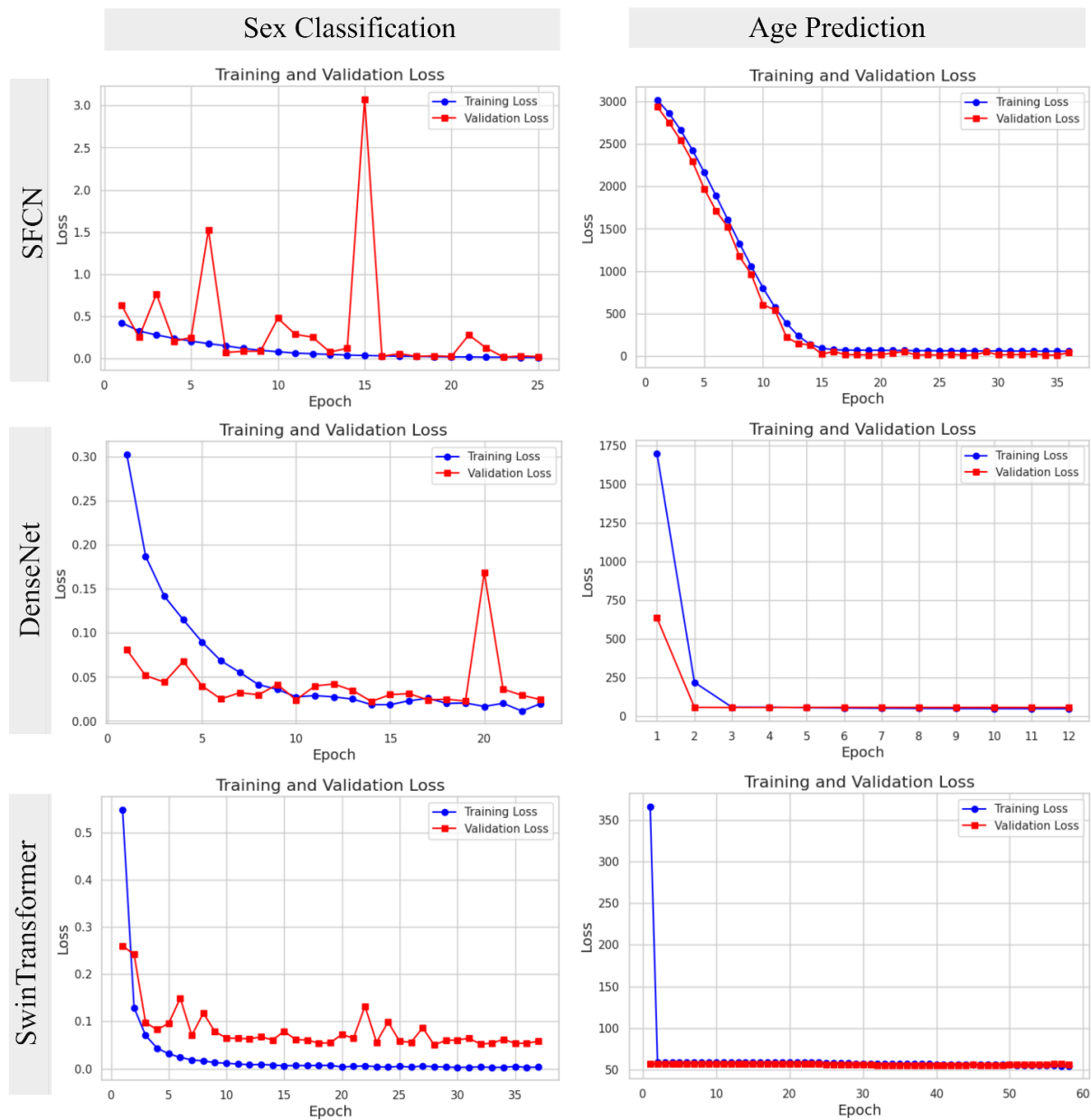

Training and Validation losses visualized for each model for all epochs until the epoch with last best loss. Early stopping was applied when the validation loss did not decrease for 10 consecutive epochs.